\documentclass[10pt,twocolumn,letterpaper]{article}

\usepackage[final]{latexdouble}      
\usepackage{amsmath}
\usepackage{mathtools} 

\usepackage{multirow}
\usepackage{pifont}
\usepackage[table]{xcolor}
\definecolor{supergray}{gray}{0.75} 
\definecolor{wakeupgreen}{HTML}{C7F4C2}

\definecolor{latexdoubleblue}{rgb}{0.21,0.49,0.74}
\usepackage[pagebackref,breaklinks,colorlinks,allcolors=latexdoubleblue]{hyperref}

\title{WakeupUrban: Unsupervised Semantic Segmentation of Mid-20$^{th}$ century Urban Landscapes with Satellite Imagery}

\author{
Tianxiang Hao$^{1*}$ \quad
Lixian Zhang$^{2*\dagger}$ \quad
Yingjia Zhang$^{3*}$ \quad 
Mengxuan Chen$^{4}$ \quad \\
Jinxiao Zhang$^{4}$ \quad 
Runmin Dong$^{5}$ \quad 
Haohuan Fu$^{1,2,4\dagger}$ \\
{\footnotesize
$^1$ Tsinghua University, SIGS $^2$ National Supercomputing Center in Shenzhen }\\
{\footnotesize
$^3$ New York University Shanghai $^4$ Tsinghua University, DESS $^5$ Sun Yat-sen University}
}

\begin{document}

\twocolumn[{%
\renewcommand\twocolumn[1][]{#1}%
\maketitle
\vspace{-0.3in} 
}]

\begingroup
\renewcommand{\thefootnote}{}
\footnotetext{$^{*}$ Equal contributions.}
\footnotetext{$^{\dagger}$ Corresponding authors.}
\endgroup

\begin{abstract} 
   Historical satellite imagery archive, such as Keyhole satellite data, offers rare insights into understanding early urban development and long-term transformation. However, severe quality degradation (\textit{e.g.}, distortion, misalignment, and spectral scarcity) and the absence of annotations have long hindered its analysis. To bridge this gap and enhance understanding of urban development, we introduce \textbf{WakeupUrbanBench}, an annotated segmentation dataset based on historical satellite imagery with the earliest observation time among all existing remote sensing (RS) datasets, along with a framework for unsupervised segmentation tasks, \textbf{WakeupUSM}. First, WakeupUrbanBench serves as a pioneer, expertly annotated dataset built on mid-$20^{\text{th}}$ century RS imagery, involving four key urban classes and spanning 4 cities across 2 continents with nearly 1000 km$^2$ area of diverse urban morphologies, and additionally introducing one present-day city. Second, WakeupUSM is a novel unsupervised semantic segmentation framework for historical RS imagery. It employs a confidence-aware alignment mechanism and focal-confidence loss based on a self-supervised learning architecture, which generates robust pseudo-labels and adaptively prioritizes prediction difficulty and label reliability to improve unsupervised segmentation on noisy historical data without manual supervision. Comprehensive experiments demonstrate WakeupUSM significantly outperforms existing unsupervised segmentation methods \textbf{both WakeupUrbanBench and public dataset}, promising to pave the way for quantitative studies of long-term urban change using modern computer vision. Our benchmark and codes will be released at \url{https://github.com/Tianxiang-Hao/WakeupUrban}.
\end{abstract}    
\section{Introduction}

\begin{figure}[ht!]
  \centering
  \includegraphics[width=\columnwidth-2.0cm]{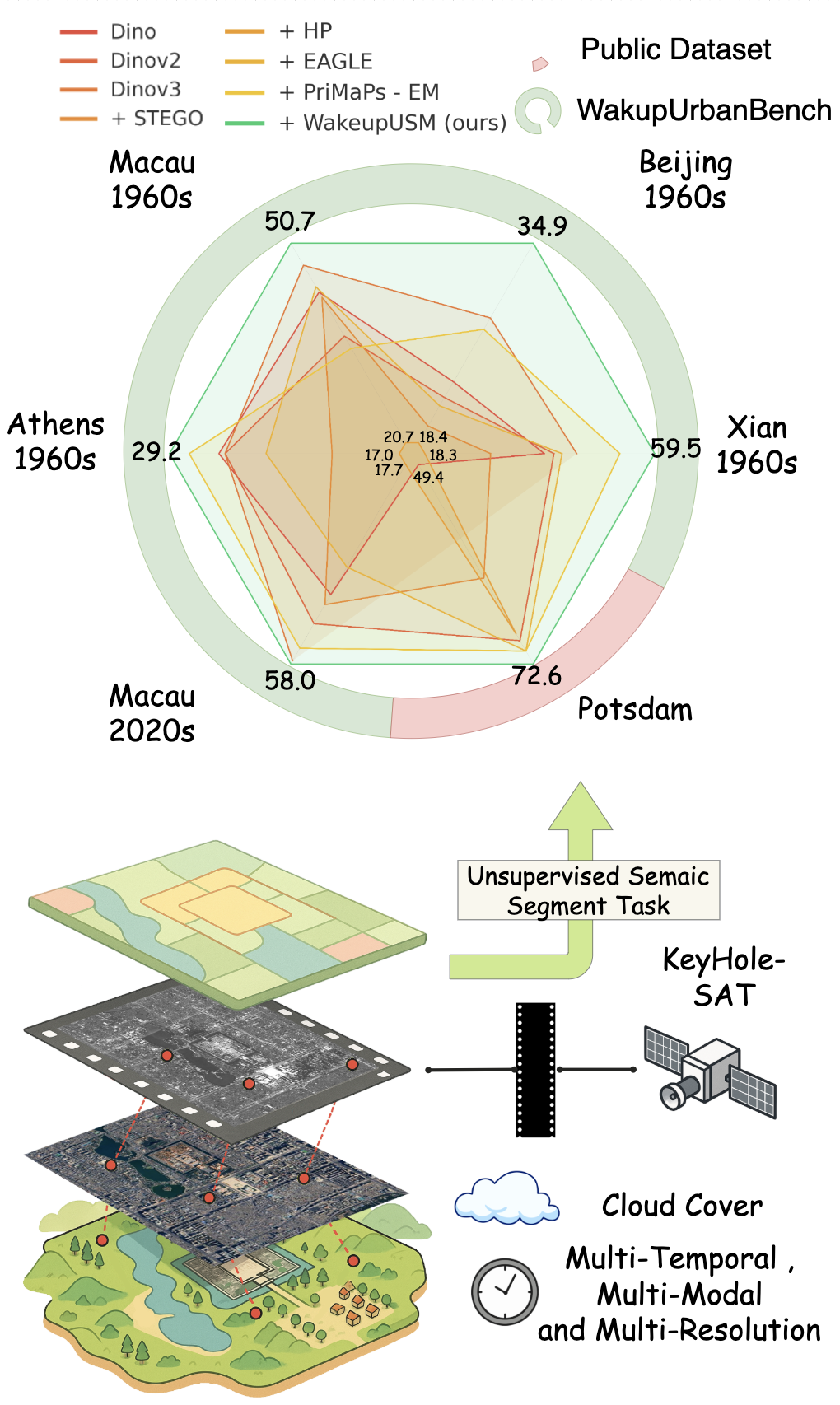}
  \caption{Overview of the WakeupUSM and WakeupUrbanBench. WakeupUrbanBench serves as a pioneer benchmark built on 1960s RS imagery, spanning multiple cities. WakeupUSM achieves superior performance on this benchmark and the public dataset, boosting interpretability in long-term urban revolutions.}
  \label{fig:WakeupUrban}
\end{figure}

The pursuit of sustainable urban development, a cornerstone of global agendas such as the Sustainable Development Goals (SDGs) \cite{UnitedNations} and broader urban sustainability initiatives \cite{yuan2020long}, necessitates a profound understanding of historical urban morphology. Analyzing past planning paradigms, infrastructure evolution, and land-use transitions in contemporary megacities provides crucial insights for addressing challenges related to urban equity, resilience, and environmental sustainability. Earth observation (EO) data from satellites offers an unparalleled opportunity to quantitatively assess urban systems across extensive spatial and temporal scales \cite{Swcare, zhang2024prolonged}.

Despite their immense potential for reconstructing long-term urban trajectories, satellite images from the mid-20$^{th}$ century remain largely underutilized. This under-exploration stems from two fundamental and interrelated challenges:

First, there is an acute lack of large-scale, meticulously curated benchmark datasets specifically designed for historical remote sensing data. Unlike modern imagery, historical archives are characterized by inherent degradations, including significant geometric distortions, limited spectral and spatial resolution, and a pronounced scarcity of ground truth annotations. These unique properties render existing contemporary benchmarks structurally incompatible \cite{Potapov2022LongTermStudies, Yu2023}. For instance, Keyhole reconnaissance data, a primary source of mid-century imagery, presents intrinsic degradations that severely impede both algorithmic interpretation and manual labeling. These include: i) a grayscale-only format, ii) spatial resolutions varying between 0.6 and 1.8 meters, and iii) severe geometric distortions, scanning artifacts, and poor contrast. Such characteristics not only obfuscate visual semantics but also render high-fidelity annotation exceptionally burdensome and ambiguous, particularly in the absence of concurrent reference data. The substantial cost and inherent ambiguity associated with large-scale manual labeling make supervised learning approaches largely infeasible, underscoring the urgent need for alternative strategies to curate and leverage these invaluable historical resources.

Second, there is a concurrent absence of scalable analytical methods specifically engineered to overcome the aforementioned data degradations. Modern deep learning approaches, while powerful, are heavily reliant on vast annotated corpora. The stark domain shift between mid-century and contemporary satellite imagery, encompassing differences in sensor modality (grayscale vs. multispectral), reflectance properties, architectural patterns, and various degradations, significantly limits the transferability of existing deep learning models. This often leads to semantic misalignment and representational collapse when applied to historical data. Consequently, in the pervasive absence of sufficient ground-truth data, conventional supervised methods become untenable. This critical gap necessitates the development of novel, unsupervised segmentation strategies capable of extracting semantics from spectrally impoverished, noisy, and distorted inputs, which is essential for enabling large-scale, quantitative analysis of early urban forms.

To address this, we introduce \textbf{WakeupUrban}, a novel framework designed for the unsupervised extraction of historically dormant urban areas from Keyhole satellite imagery. As presented in Fig. \ref{fig:WakeupUrban}, WakeupUrban consists of: (i) \textbf{WakeupUrbanBench}, the earliest-time semantic segmentation dataset among all existing remote sensing benchmarks, and (ii) \textbf{WakeupUSM}, a robust model for unsupervised segmentation.

WakeupUrbanBench is the first professionally annotated semantic segmentation dataset derived from declassified mid-20th-century Keyhole satellite imagery. It spans Macau, Athens, Beijing, and Xi’an—covering two continents and four cities—and provides carefully curated multi-class land-use labels and a binary impervious-surface product. Built through expert interpretation and extensive cross-referencing of archival sources, the dataset represents a substantial technical and logistical advance, enabling quantitative analyses of historical urban development that were previously constrained by data scarcity.

To fully leverage WakeupUrbanBench, the WakeupUSM is proposed as a novel unsupervised segmentation method engineered to overcome the methodological barriers associated with historical satellite imagery. WakeupUSM integrates the Segment Anything Model (SAM) for initial pseudo-label generation with a dual-branch DINO-based self-supervised learning architecture. The key contributions include a confidence-aware alignment mechanism that transforms diverse zero-shot masks from the SAM into reliable pseudo-labels and Focal-Confidence Loss that integrates class-level prediction difficulty and pixel-level label reliability to precisely focus learning on trustworthy and semantically challenging regions, leading to significantly improved unsupervised segmentation of historical imagery without additional manual annotation.

In summary, our contributions are threefold: 1) We introduce \textbf{WakeupUrbanBench}, the first semantic segmentation dataset covering mid-20th-century grayscale satellite imagery, enabling urban land use and impervious surface analysis under grayscale degradation. 2) We propose \textbf{WakeupUSM}, a novel framework for unsupervised segmentation and classification of historical satellite imagery that reduces the dependence on manual annotation. 3) Experimental results demonstrate that WakeupUSM outperforms five DINO-based self-supervised approaches and several classical unsupervised baselines, underscoring its effectiveness in high-resolution Keyhole RS image segmentation and its potential to enhance interpretability in historical remote sensing analysis.

\section{Related Work}
\begin{figure}[htbp]
    \centering
    \includegraphics[width=\linewidth]{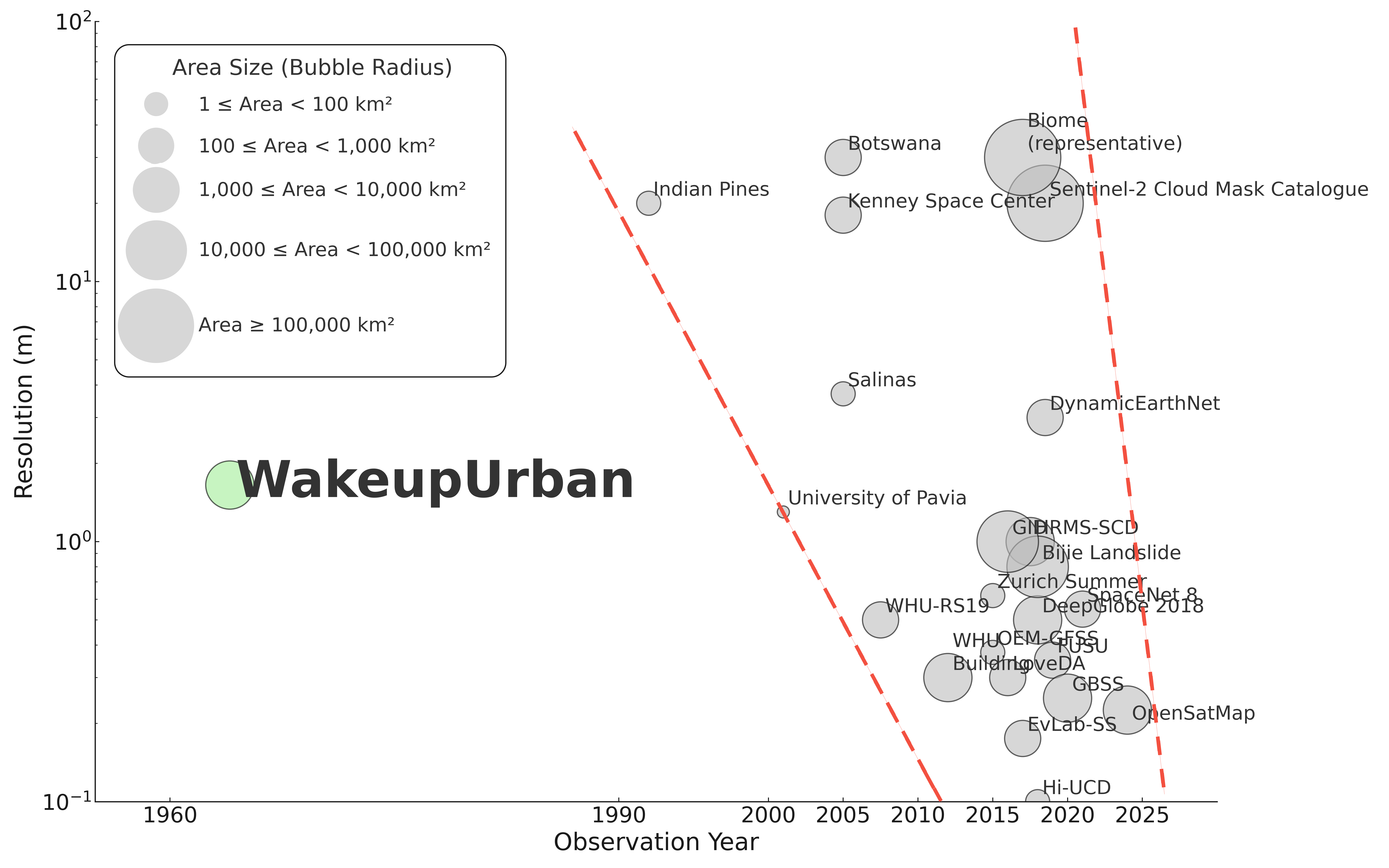}
    \caption{Visualization of high-resolution remote sensing segmentation datasets by year, resolution, and coverage. More recent datasets show greater diversity in resolution and extent, while WakeupUrbanBench is the earliest with relatively high resolution, bridging a key temporal gap.}
    \label{fig:comparison}
\end{figure}

\subsection{Segmentation datasets in Remote Sensing}

Segmentation datasets are fundamental for advancing remote sensing applications. We systematically review the existing datasets in remote sensing segmentation (detailed in supplementary\cite{OpenSatMap2024,OEM_GFSS2024,FUSU2024,GBSS2024,OpenEarthMap2022,Hansch_2022_CVPRW,gic_hyperspectral_datasets,95-cloud,Tian2020HiUCD,HRMS_SCD2025,francis2020sentinel2,Toker_2022_CVPR,LeSaux2018DataFusion,Demir_2018_CVPR_Workshops,Ji2020Landslide,Russo2024SEN12WATER,Saha2017Multimodal,Foga2017CloudDetection,Zhang2017Learning,USGS2016SPARCS,Hughes2014CloudDetection,Wang2021LoveDA,Volpi_2015_CVPR_Workshops,xia2010structural,dai2011satellite,GID2020,Ji2019FCNBuildingExtraction}). It is clear to observe a distinct and problematic evolutionary trajectory that while recent years have witnessed a proliferation of datasets characterized by impressive diversity in spatial resolutions and expansive geographic coverage, a fundamental limitation persists in their temporal depth. As depicted in Fig. \ref{fig:comparison}, the prevailing trend strongly favors contemporary imagery, often capturing sub-meter to few-meter details of present-day land cover. Conversely, datasets extending into earlier periods, particularly those preceding the satellite era's mature stages (\textit{e.g.}, 1990s - 2010s), exhibit significantly coarser spatial resolutions and constrained geographic extents. This disparity largely stems from historical sensor capabilities, the complexities of analog data acquisition, and the challenges of large-scale digitization and georeferencing. This pronounced historical data gap \cite{Potapov2022LongTermStudies, Yu2023} critically impedes granular, long-term urban change analysis, which is vital for understanding historical development trajectories, assessing the efficacy of past planning policies, and informing future sustainability strategies. Addressing this temporal discontinuity is therefore paramount for a holistic understanding of global urban evolution.

The WakeupUrbanBench dataset directly addresses this critical gap. The WakeupUrbanBench is the first pixel-level segmentation dataset derived from high-resolution, 1960s grayscale imagery, covering an expansive over 1,000 km$^2$. Beyond its inherent value for remote sensing and social science, WakeupUrbanBench provides historical high-resolution imagery with comprehensive multi-label annotations to support land-use changes and sustainable development research, establishing a crucial benchmark for evaluating unsupervised semantic segmentation models in historically degraded contexts. 

\subsection{Segmentation model in Remote Sensing}
Remote sensing image segmentation often has several challenging features such as severe imbalance in foreground and background distribution, complex background, intra-class heterogeneity, inter-class homogeneity and tiny objects\cite{AerialFormer}. These challenges are intensified in our task due to grayscale inputs, variable resolution, and historical image degradation. Most segmentation models use CNN or Transformer-based architectures. A method is proposed for the single-target segmentation task, where the existing segmentation performance of Large SAM is leveraged by freezing its parameters, while a Small model generates boxes prompts, to assist SAM in achieving accurate segmentation of urban villages from satellite images \cite{UV-SAM}. For the semantic segmentation task of remote sensing images, a simple and effective framework is proposed that uses the objects and boundaries generated by the SAM and assists model training by designing object consistency loss and boundary preservation loss\cite{SAM-Assisted}.
\begin{figure*}[htbp]
  \centering
  \includegraphics[width=\linewidth]{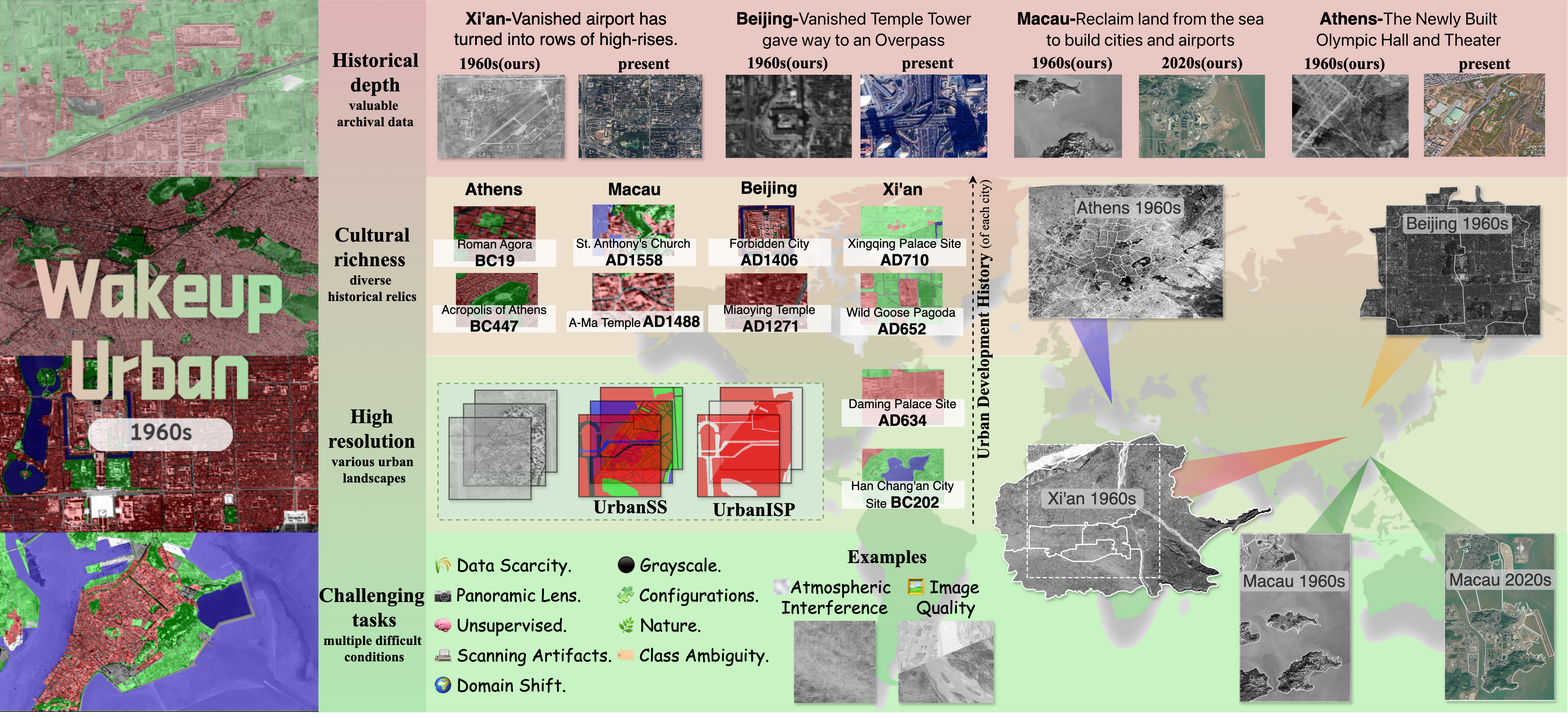} 
  \caption{Overview of WakeupUrbanBench, detailing its spatial structure, inherent task-specific challenges, semantic category design, and significance for historical cross-era urban dynamics research.}
  \label{fig:dataset}
\end{figure*}

\subsection{Unsupervised semantic segmentation in computer vision}

Semantic segmentation based on deep learning often utilizes encoder-decoder architectures for pixel-wise classification \cite{zhou2024image}. The existing researchers have explored various prompt mechanisms by designing unique decoders equipped with multiple prompt recognition features and task-specific functions, achieving notable performance in zero-shot semantic segmentation \cite{SEEM}\cite{xdecoder}\cite{OpenSEED}. From an unsupervised learning perspective, a common approach involves two key steps: the generation of pseudo masks and the subsequent training of models using these automatically derived labels. For example, STEGO \cite{STEGO} proposes a method that combines clustering and Conditional Random Field (CRF) refinement to perform knowledge distillation by leveraging the inherent correlations within self-supervised features. Another approach, HP \cite{HP}, contributes to mine global and local hidden positive samples, achieving performance improvements through contrastive learning. Additionally, Primaps \cite{Primaps} utilizes the intrinsic properties of self-supervised learning features to construct pseudo labels, ultimately leading to unsupervised semantic segmentation. However, most methods fail under grayscale, heavily degraded data as historical RS images. WakeupUSM addresses this gap with a framework tailored to such conditions.

\section{WakeupUrbanBench Dataset}

WakeupUrbanBench is a high-fidelity, city-scale semantic segmentation benchmark built from declassified 1960s Keyhole panchromatic reconnaissance imagery \cite{keyhole2007repository} co-registered with contemporaneous official urban maps. We provide pixel-level labels for impervious surface and five land-use categories—roads, buildings, water bodies, green lands, and other urban types. Spanning four cities across two countries and two periods (1960s and 2020s), the benchmark constitutes a challenging domain due to grayscale distortions and limited spatial resolution in reconnaissance imagery, and it enables studies in historical urban morphology, cultural-heritage conservation, and cross-temporal/cross-city generalization. All assets (original imagery, semantic masks, impervious-surface “ISP” masks, metadata, and baseline scripts), together with detailed documentation, will be publicly released under a Creative Commons Attribution–NonCommercial (CC BY-NC) license on GitHub and Hugging Face.

\subsection{Dataset Construction}

\textbf{Data Areas}: The 1960s collections capture distinctive urban morphologies—\emph{Xi’an} with its orthogonal walled core and ceremonial axes; \emph{Beijing} with an imperial grid of hutong/siheyuan transitioning toward early ring-road expansion; \emph{Athens} with a classical core embedded in a topography-molded irregular network; and \emph{Macau (1960s)} presents a dense waterfront urban fabric shaped by the shoreline and land-reclamation edges. To enable cross-temporal generalization and change detection, we additionally include a contemporary \emph{Macau (2020s)} subset that allows direct comparison across several decades of intense urban development and land reclamation.

\noindent\textbf{Data Acquisition}: Remote sensing imagery was acquired in 1960s by the U.S. Keyhole reconnaissance satellite and declassified in recent decades, equipped with a panoramic camera featuring a focal length with an effective ground sampling distance (GSD) of approximately 0.6 m/pixel in the city center and approximately 2.7 m/pixel in the sub-urban area, capturing panchromatic film at nadir with film digitization at 12 bits depth.

\noindent\textbf{Radiometric and Geometric Preprocessing}: Original film scans underwent histogram matching, control-point-based geo-referencing, Rational Polynomial Coefficients (RPC) refinement for abnormal brightness reduction, fine geospatial registration, relief displacement correction, and contrast variations reduction across overlapping frames. Please refer to the supplementary material for more details.

\noindent\textbf{Annotation}: Subsequent to radiometric and geometric preprocessing, remote sensing specialists annotated the dataset using comparative analysis between Keyhole satellite imagery and the 1965 urban map derived from the Department of Shaanxi Provincial Archives. Remote sensing specialists with high historical imagery interpretation experience performed manual annotation with a final pass to ensure topological consistency and verify that major landmarks (\textit{e.g.}, cultural heritage) align correctly. Annotations are produced in two formats: impervious surface product (ISP) and urban land use classification. The latter adheres to standard remote sensing classification categories\cite{2013isprs}, including roads and transportation facilities, buildings, water bodies, green spaces, farmland, and other urban-use types, serving as a valuable reference for urban development analysis in China and globally during the 1960s.

\subsection{Dataset Exploration}

\begin{figure*}[htbp]
  \centering
  \includegraphics[width=\linewidth]{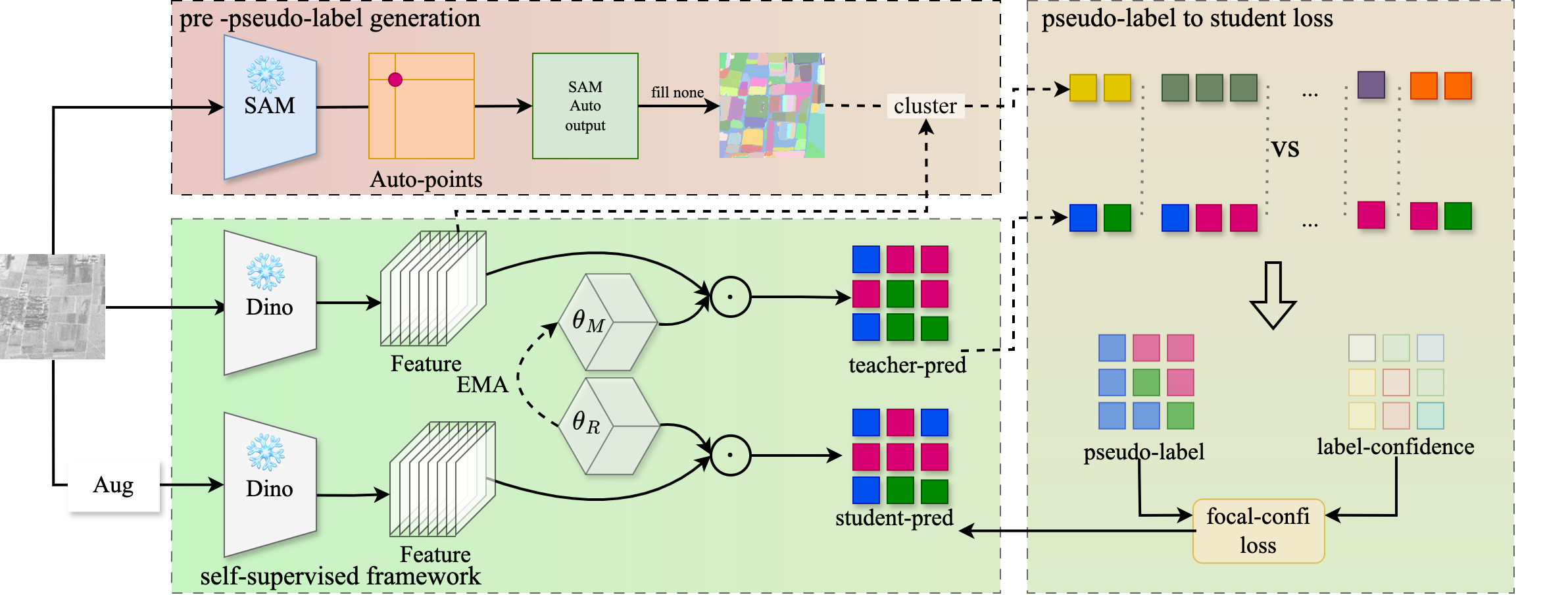} 
  \caption{Architecture of WakeupUSM, an unsupervised segmentation model designed for urban scene analysis. WakeupUSM integrates a self-supervised learning (SSL) backbone under three stages: 1) Pre-Pseudo-Label Generation, 2) Self-Supervised Framework, and 3) Pseudo-Label to Student Loss.}
  \label{fig:method}
\end{figure*}
\textbf{Overview}:
As shown in Fig. \ref{fig:dataset}, WakeupUrbanBench is currently a city-level urban landcover segmentation dataset for demonstration. The dataset consists of two products: (1)\textbf{UrbanISP-1960s}: the WakeupUrbanISP impervious surface product; (2)\textbf{UrbanSS-1960s}: the WakeupUrbanSS semantic segmentation product for urban landcover. The UrbanISP-1960s and UrbanSS-1960s subsets are divided into training, validation, and testing sets, consisting of 614, 154, and 256 image-label pairs, respectively.

\noindent\textbf{Hignlights}:
We summarize the key features of WakeupUrbanBench as follows:
(1) \textbf{Pioneering Benchmark}: WakeupUrbanBench is the first segmentation benchmark built on 1960s Keyhole imagery, enabling historical urban analysis and zero-shot generalization across cities and decades, promoting advancements in both remote sensing (RS) and computer vision (CV) communities. (2) \textbf{Challenging Conditions}: WakeupUrbanBench contains the inherent complexities of historical remote sensing imagery, including grayscale imaging, geometric distortions, and resolution limitations. These challenges demand models with strong generalization in unsupervised settings, making WakeupUrbanBench more difficult than most modern benchmarks. (3) \textbf{Comprehensive Urban Landscape Coverage}: WakeupUrbanBench captures diverse urban features and land cover types in a historically significant Chinese city. It includes detailed urban elements such as roads, buildings, water bodies, green areas, and farmland, all of which are manually annotated by domain experts. This provides a valuable resource for the analysis of historical urban form, planning trends, and socio-economic landscapes within the broader context of global urban studies.

\section{WakeupUSM}


To achieve the unsupervised segmentation on WakeupUrbanBench, the core intuition is to synergistically combine the class-agnostic structural boundaries and the rich semantic groupings. However, any pseudo-labels from this initial fusion are inevitably imperfect and reflect the data's inherent ambiguity. Therefore, WakeupUSM is designed to critically manage their unreliability. As shown in Fig. \ref{fig:method}, WakeupUSM consists of three stages: pre-pseudo-label generation, self-supervised learning, and loss computation. Initially, the SAM generates region-level masks from unlabeled images via grid-based prompts, which are aggregated into complete pseudo-label maps. In parallel, a frozen DINO-based dual-branch network extracts patch-level features, with the teacher updated via EMA. SAM and teacher outputs are compared with the student’s predictions to compute consistency-aware supervision and a confidence map, which are fused via a novel focal-confidence loss. This loss adaptively emphasizes reliable and semantically challenging regions to manage the noisy supervision signal.

\subsection{Pre-Pseudo-Label Generation and Self-Supervised Framework}

\paragraph{Stage I: Pre-pseudo-label generation.}
Given an input image $x \in \mathbb{R}^{H \times W \times 3}$, we first obtain structure-aware binary proposals using the Segment Anything Model (SAM) under point prompts.
Prompt locations $\{p_n\}_{n=1}^{N}$ are sampled by a Poisson-disk or jittered-grid strategy to balance spatial coverage and redundancy.
For each prompt $p_n$, SAM produces a salient binary mask $B_n \in \{0,1\}^{H \times W}$.
The set $\{B_n\}$ is merged into an initial canvas by (i) confidence/size filtering to remove degenerate regions, (ii) conflict-aware compositing with an overlap rule (e.g., IoU threshold) in descending saliency, and (iii) a lightweight morphological closing to bridge small gaps.
To address uncovered areas, we perform a \emph{coverage filling} step that assigns labels to previously unmarked pixels by nearest-region propagation (e.g., geodesic distance to the merged foreground) accompanied by a small region-growing pass.
For elongated and over-merged structures (e.g., roads or rivers), a \emph{skeleton-based width division} refines shapes: we compute a morphological skeleton and local radius (via distance transform) and reconstruct thin, topology-preserving regions by expanding from the skeleton with locally adaptive width.

In parallel, we extract per-pixel (or patch-level reprojected) features $\phi(i)$ using a \textbf{frozen} dual-branch DINO backbone (no training, no EMA updates).
Given the predefined number of classes $C$, we run KMeans with $K{=}C$ on $\{\phi(i)\}$ to obtain a cluster assignment map $Q \in \{1,\dots,C\}^{H \times W}$.
We then \emph{fuse} SAM-derived regions and clustered semantics to form categorical pseudo-labels:
within each merged SAM region, the final label is set by majority voting over $Q$ inside the region; pixels outside SAM coverage inherit their labels directly from $Q$.
This yields a full-resolution, pixel-wise prior
\[
P \in \{1,\dots,C\}^{H \times W},
\]
which serves as the \emph{pre-pseudo-label} for downstream learning.
No per-pixel weights are used at this stage.

\paragraph{Stage II: Self-supervised architecture.}
We adopt the dual-branch DINO architecture and the lightweight segmentation head module from Primaps-EM~\cite{Primaps}.
Two augmented views $x^{(1)}$ and $x^{(2)}$ of the same image are forwarded through the fixed teacher and student encoders to produce patch tokens and their projections in a shared embedding space.
These features are then consumed by a lightweight segmentation head to generate dense semantic predictions.
The optimization details (e.g., objectives and schedules) are deferred to a later section; here, the backbone parameters (teacher/student encoders and projectors) remain unchanged throughout, serving purely as a stable feature extractor.
The Stage I prior $P$ provides structural and categorical guidance for this stage.

\subsection{Pseudo-Label and Student Loss}

\paragraph{Overview.}
The student is supervised using pseudo-labels derived from SAM regions produced in Stage~I.
All features/backbones and the teacher branch remain frozen; the objective in this subsection updates the student only.
To convert regional evidence into pixel-wise training targets without introducing an additional ignore state, we employ a
\emph{confidence-aware alignment mechanism} that adaptively selects hard (one-hot) versus soft (distributional) targets.

\paragraph{Confidence-aware alignment mechanism.}
Let $K$ be the number of classes and let $\mathcal{R}$ denote a merged SAM region.
Define the empirical class proportions within $\mathcal{R}$ by
\[
p_c(\mathcal{R}) \;=\; \frac{1}{|\mathcal{R}|}\sum_{i\in\mathcal{R}}\mathbf{1}\{P_i=c\},\qquad c\in\{1,\dots,K\},
\]
where $P_i$ is the Stage~I pseudo-label at pixel $i$.
Let the dominant class be $c^*=\arg\max_c p_c(\mathcal{R})$ with peak proportion $p_{c^*}(\mathcal{R})$.

\textbf{Threshold.}
Given an alignment strength $B>0$, we set the high-confidence threshold to
\[
\tau_{\mathrm{high}} \;=\; \frac{B}{\,B + K - 1\,}.
\]
This expression admits a Bayesian interpretation: it corresponds to a Dirichlet prior whose pseudo-counts place $B$ on the dominant hypothesis and $1$ on all others, yielding a prior mass of $B/(B{+}K{-}1)$ for the mode.
Hence, larger $B$ enforces stricter confidence (\,$\tau_{\mathrm{high}}\!\to\!1$ as $B\!\to\!\infty$\,), while larger $K$ relaxes the requirement for the same $B$.

\paragraph{Region-to-pixel alignment and targets.}
Using the regional class proportions $p_c(\mathcal{R})$ and the threshold $\tau_{\mathrm{high}}$ defined above, we treat regions with $p_{c^*}(\mathcal{R}) \ge \tau_{\mathrm{high}}$ as high-confidence and assign them a hard label $c^*$, while the remaining regions keep the full distribution $p(\mathcal{R}) = \{p_c(\mathcal{R})\}_{c=1}^{K}$ as a soft target; in both cases the regional confidence is $\text{Conf}(\mathcal{R}) = p_{c^*}(\mathcal{R})$.
For training, we require pixel- or patch-level targets on the student prediction grid $\mathcal{G}$.
For any spatial element $u\in\mathcal{G}$, let $\mathcal{R}(u)$ be the SAM region covering $u$ (if a rare overlap occurs, we pick the region with larger area or larger $p_{c^*}$), and define

\begin{equation}
y_u=
\begin{cases}
\mathrm{one\_hot}(c^*), & p_{c^*}(\mathcal{R}(u)) \ge \tau_{\mathrm{high}},\\[2pt]
p(\mathcal{R}(u)), & \text{otherwise,}
\end{cases}
\label{eq:conf-align}
\end{equation}
\begin{equation}
c_u = p_{c^*}(\mathcal{R}(u)).
\end{equation}
When $\mathcal{G}$ is a patch grid, we map pixels to patches by majority pooling in the hard case and average pooling in the soft case.
To reduce boundary artifacts, we optionally apply a one-pixel band of morphological erosion before querying $\mathcal{R}(u)$ and fall back to nearest-region assignment for uncovered elements.

\paragraph{Properties and practical choices.}
(i) \emph{Scale-invariance in region size}: the use of proportions $p_c(\mathcal{R})$ makes decisions insensitive to $|\mathcal{R}|$.
(ii) \emph{Controllability}: $B$ directly tunes the strictness of hard-label adoption; typical values $B\in[1,5]$ work robustly across datasets.
(iii) \emph{Class-multiplicity effect}: for fixed $B$, larger $K$ lowers $\tau_{\mathrm{high}}$, preventing an overly conservative policy when many categories coexist.
(iv) \emph{No ignore state}: all spatial elements receive either a hard one-hot or a soft distributional target, ensuring dense supervision.

\paragraph{Focal-Confidence loss.}
Let $x_u$ be the student's logits at $u\in\mathcal{G}$ and let $\mathrm{CE}(x_u,y_u)$ denote cross-entropy with $y_u$ (hard or soft).
To emphasize rare/difficult classes, we employ a class-difficulty factor $(1-\bar{p}_{y_u})^\gamma$, where $\bar{p}_{y_u}$ is the batch-average predicted probability of the target; for soft targets, $\bar{p}_{y_u}=\sum_c y_{u,c}\,\bar{p}_c$ with $\bar{p}_c$ the batch mean for class $c$.
With per-element confidence $c_u=p_{c^*}(\mathcal{R}(u))$, the objective is
\begin{equation}
\mathcal{L}_{\text{focal-confi}} 
= \frac{1}{|\mathcal{G}|} \sum_{u\in\mathcal{G}}
\Big[
\alpha \cdot (1-\bar{p}_{y_u})^{\gamma} \cdot \mathrm{CE}(x_u, y_u) \cdot c_u^{\beta}
\Big],
\label{eq:focal-confi}
\end{equation}

where $\gamma,\beta$ control focusing strength (e.g., $\gamma{=}2,\ \beta{=}0.5$) and $\alpha$ balances the overall scale.
This loss, together with the confidence-aware alignment mechanism, provides dense supervision for the student while all feature extractors remain fixed.

\paragraph{Implementation notes.}
We precompute $p_c(\mathcal{R})$ on the pixel grid from Stage~I labels and cache region-wise tuples
$\big(c^*,\,p_{c^*}(\mathcal{R}),\,p(\mathcal{R})\big)$.
During training, targets are assembled on the fly by table lookups using region ids; the added overhead is negligible compared with the forward pass.

\section{Experiments}

\begin{table*}[t]
\centering
\caption{Comparison of unsupervised segmentation performance across datasets, reported as mIoU/Acc (\%). For each method and dataset, we select the best-performing backbone size. \textbf{Bold} denotes the best within each method group, and \textbf{\textcolor{red}{Red}} marks the overall best.}
\label{experiments-multidatasets-unsupervised}
\resizebox{\linewidth}{!}{
\begin{tabular}{l|c|cc|cc|cc|cc|cc}
\hline
\multirow{2}{*}{Method} & \multirow{2}{*}{Backbone}
& \multicolumn{2}{c|}{Xi'an1960s}
& \multicolumn{2}{c|}{Beijing1960s}
& \multicolumn{2}{c|}{Macow1960s}
& \multicolumn{2}{c|}{Athens1960s}
& \multicolumn{2}{c}{Macow 2020s} \\
\cline{3-12}
& & mIoU & Acc & mIoU & Acc & mIoU & Acc & mIoU & Acc & mIoU & Acc \\
\hline
\textit{Zero-shot} &&&&&&&&&\\
Grounded SAM\cite{groundedSAM} & SwinT–ViT-H & 10.19 & 13.25 & 22.44 & 65.40 & 14.87 & 15.63 & \textbf{15.35} & \textbf{46.96}&21.61&36.33 \\
XDecoder(T)\cite{xdecoder} &  FocalT& 11.82 & 23.11 & 18.64 & \textbf{67.85} & 11.86 & 22.17 & 14.75 & 37.86 &31.62 &48.60\\
XDecoder(L)\cite{xdecoder} & FocalL & 12.74 & 11.91 & 14.59 & 46.24 & 13.22 & 23.77 & 6.27 & 21.42 &49.86& 72.77\\
SEEM(T)\cite{SEEM} & FocalT & \textbf{12.90} & 18.04 & \textbf{24.94} & 57.72 & \textbf{23.64} & 29.54 & 14.92 & 42.00&49.10 &75.48 \\
SEEM(L)\cite{SEEM} & FocalL & 9.28 & 16.72 & 10.40 & 30.74 & 13.81 & 20.93 & 10.37 & 24.67&45.71& 66.15 \\
\hline
Dino\cite{Dino} & DINO ViT - /8
& 39.68 & 74.09 & 23.46 & 67.05 & 43.32 & 90.45 & 26.55 & 64.24 & 43.98 & 69.77 \\
+ STEGO \cite{STEGO} & DINO ViT - /8
& 30.06 & 52.88 & 19.80 & 44.94 & 42.56 & \textbf{\textcolor{red}{92.64}} & 20.57 & 44.64 & 46.02 & 65.39 \\
+ HP \cite{HP} & DINO ViT - /8
& 18.27 & 72.50 & 18.43 & \textbf{71.11} & 20.74 & 75.28 & 17.02 & 60.80 & 17.71 & 49.17 \\
+ EAGLE \cite{EAGLE} & DINO ViT - /8
& 42.86 & 68.41 & 21.47 & 42.12 & \textbf{44.12} & 91.56 & 24.06 & 50.77 & 38.39 & 54.16 \\
+ PriMaPs - EM \cite{Primaps} & DINO ViT - /8
& 53.22 & 74.47 & 25.40 & 69.46 & 34.89 & 74.52 & 24.93 & 59.75 & \textbf{54.84} & 81.78 \\
\rowcolor{wakeupgreen}+ WakeupUSM (ours) & DINO ViT - /8
& \textbf{\textcolor{red}{59.46}} & \textbf{\textcolor{red}{88.30}} & \textbf{34.91} & 59.81 & 40.30 & 85.23 & \textbf{29.00} & \textbf{\textcolor{red}{68.51}} & 53.28 & \textbf{83.12} \\
\hline
Dinov2\cite{Dinov2} & DINOv2 ViT - /14
& 41.41 & \textbf{79.11} & 22.16 & 39.23 & 36.72 & 75.46 & 26.22 & 61.03 & \textbf{49.87} & \textbf{71.80} \\
+ PriMaPs - EM \cite{Primaps} & DINOv2 ViT - /14
& 36.30 & 71.61 & \textbf{27.81} & \textbf{62.73} & 33.88 & 64.78 & 28.12 & 65.87 & 48.34 & 68.26 \\
\rowcolor{wakeupgreen}+ WakeupUSM (ours) & DINOv2 ViT - /14
& \textbf{43.01} & 66.24 & 26.98 & 59.62 & \textbf{37.78} & \textbf{75.73} & \textbf{\textcolor{red}{29.16}} & \textbf{66.72} & 49.10 & 68.80 \\
\hline
Dinov3\cite{Dinov3} (web) & DINOv3 ViT - /16 (web)
& 45.50 & 79.60 & 28.76 & 73.78 & 36.69 & 65.71 & 26.23 & 63.66 & 57.37 & 82.02 \\
\rowcolor{wakeupgreen}+ WakeupUSM (ours) & DINOv3 ViT - /16 (web)
& \textbf{46.82} & \textbf{82.06} & \textbf{\textcolor{red}{34.94}} & \textbf{\textcolor{red}{73.85}} & \textbf{40.49} & \textbf{75.01} & \textbf{26.70} & \textbf{64.91} & \textbf{\textcolor{red}{58.00}} & \textbf{\textcolor{red}{84.66}} \\
\hline
Dinov3\cite{Dinov3} (sat) & DINOv3 ViT - /16 (sat)
& \textbf{35.50} & 64.86 & \textbf{24.53} & 55.60 & 47.36 & 90.30 & 22.18 & 47.76 & \textbf{44.69} & 61.40 \\
\rowcolor{wakeupgreen}+ WakeupUSM (ours) & DINOv3 ViT - /16 (sat)
& 33.28 & \textbf{66.02} & 23.00 & \textbf{56.54} & \textbf{\textcolor{red}{50.66}} & \textbf{92.15} & \textbf{23.89} & \textbf{57.29} & 44.63 & \textbf{61.94} \\
\hline
\end{tabular}}
\end{table*}

\begin{table}[t]
\centering
\caption{Semantic segmentation results on Potsdam-3 only. We keep the original formatting and, for \textbf{Dino} and \textbf{DINOv2} families, merge different backbones by selecting the better-performing one for display.}
\resizebox{\linewidth}{!}{
\begin{tabular}{l|c|cc}
\hline
\multirow{2}{*}{Method} & \multirow{2}{*}{Backbone}
& \multicolumn{2}{c}{Potsdam-3} \\
\cline{3-4}
& & mIoU & Acc \\
\hline
IIC \cite{IIC} & ResNet18
& - & 65.1 \\
\hline
Dino \cite{Dino} & DINO ViT - /8
& 49.4 & 66.1 \\
+ STEGO \cite{STEGO} & DINO ViT - /8
& 62.6 & 77.0 \\
+ HP \cite{HP} & DINO ViT - /8
& 69.1 & 82.4 \\
+ EAGLE \cite{HP} & DINO ViT - /8
& 71.1 & 83.3 \\
+ PriMaPs - EM \cite{Primaps} & DINO ViT - /8
& 69.1 & 82.4 \\
\rowcolor{wakeupgreen}+ WakeupUSM (ours) & DINO ViT - /8
& 68.9 & 81.1 \\
\hline
Dinov2 \cite{Dinov2} & DINOv2 ViT - /14
& 69.9 & 82.4 \\
+ PriMaPs - EM \cite{Primaps} & DINOv2 ViT - /14
& 71.1 & 83.2 \\
\rowcolor{wakeupgreen}+ WakeupUSM (ours) & DINOv2 ViT - /14
& \textbf{\textcolor{red}{72.6}} & \textbf{\textcolor{red}{84.2}} \\
\hline
\end{tabular}
}
\label{experiments-potsdam-only}
\end{table}

\begin{table}[t]
\centering
\caption{Ablation study of \textbf{WakeupUSM} on Xi'an 1960s.}
\begin{tabular}{l|cc}
\hline
Variant & mIoU & Acc \\
\hline
w/o $\tau_{\mathrm{high}}$          & 53.97 & 80.32 \\
w/o confidence-weighted        & 45.06 & 70.33 \\
w/o CRF                             & 57.25 & 86.82 \\
\rowcolor{wakeupgreen}\textbf{WakeupUSM (full)} & \textbf{59.46} & \textbf{88.30} \\
\hline
\end{tabular}
\label{tab:wakeup-ablation}
\end{table}

\subsection{Experimental Setups}

\paragraph{Selected competing methods.}
We compare WakeupUSM with three categories of representative baselines:
(1) clustering-only methods that rely on DINO ~\cite{Dino} / DINOv2 ~\cite{Dinov2} / DINOv3 ~\cite{Dinov3} features;
(2) self-supervised semantic segmentation frameworks built upon DINO/DINOv2, including STEGO ~\cite{STEGO}, HP ~\cite{HP}, EAGLE ~\cite{EAGLE}, and PriMaPs-EM ~\cite{Primaps};
(3) zero-shot foundation models and promptable segmentors evaluated out-of-the-box, including Grounded-SAM ~\cite{groundedSAM}, XDecoder (T/L) ~\cite{xdecoder}, and SEEM (T/L) ~\cite{SEEM}. \textit{Refer to the supplementary material for more implementation details.}

\paragraph{Evaluation metrics.}
Following standard practice for unsupervised segmentation, we report mean Intersection over Union (mIoU) and Overall Accuracy (Acc).

\subsection{Comparison Results}

As summarized in Tab.~\ref{experiments-multidatasets-unsupervised} (\textit{with extended results in the supplementary material}), WakeupUSM demonstrates state-of-the-art performance in unsupervised settings for cross-historical city dataset segmentation tasks. The experimental results confirm the method's high robustness when facing complex challenges such as grayscale inputs, variable ground sampling distances (GSD, e.g., $0.6$--$2.7$,m), and scanning artifacts.

Specifically, when using DINO ViT-Base/8 as the backbone, our method achieves exceptional performance on the \emph{Xi'an1960s} dataset, reaching \textbf{59.46\% mIoU / 88.30\% Acc}. This represents a significant improvement of {+6.24\% mIoU / +13.83\% Acc} over the strongest baseline (PriMaPs-EM). Concurrently, on the \emph{Athens1960s} dataset, our method also attains \textbf{29.00\% mIoU / 68.51\% Acc}, surpassing existing methods on both key metrics. On the \emph{Macow 2020s} dataset, WakeupUSM achieves 83.12\% Acc (outperforming PriMaPs-EM's 81.78\%) while maintaining a competitive mIoU level.

To validate the model's generalization capabilities, we further evaluated the DINOv2 ViT-Large/14 backbone. The results show that WakeupUSM remains highly competitive, setting new mIoU records on \emph{Macow1960s} (37.78\% vs. 36.72\%) and \emph{Athens1960s} (29.16\% vs. 28.12\%), while sustaining comparable accuracy levels. When adopting the DINOv3 backbone, the performance gains exhibit consistency on both \textit{web} and \textit{sat} variants. The comprehensive evaluation results robustly demonstrate that the proposed WakeupUSM is effective and scalable in consistently validating across different backbones and configurations.

\subsection{Evaluation on Public Dataset}

To extend the validation of our unsupervised image segmentation framework beyond our custom benchmark, we incorporate evaluation on a well-established public dataset, \emph{Potsdam-3} \cite{wang2022unetformer}. This dataset consists of high-resolution aerial RGB imagery (38 tiles of 6000$\times$6000 pixels at 5 cm ground sampling distance) annotated with three primary classes: impervious surfaces, buildings, and low vegetation. By assessing our method on Potsdam-3, we aim to demonstrate its generalizability and robustness across diverse data distributions, bridging the gap between domain-specific applications and standardized remote sensing tasks. This evaluation aligns with established protocols from baselines such as STEGO, HP, EAGLE, and PriMaPs-EM, utilizing DINO and DINOv2 backbones.

Our WakeupUSM framework achieves superior performance in terms of mIoU and accuracy on Potsdam-3 under equivalent backbone configurations, underscoring its efficacy in handling standard high-resolution RS imagery without supervision. These results affirm that the adaptive mechanisms in WakeupUSM, such as dynamic feature awakening and multi-scale integration—translate effectively to public benchmarks, potentially offering a versatile foundation for broader applications of urban monitoring. \textit{Refer to the supplementary material for comprehensive results on other public datasets.}


\subsection{Ablations and Qualitative Evidence}

As reported in Tab.~\ref{tab:wakeup-ablation} on \emph{Xi’an 1960s}, the removal of confidence-weighted supervision causes the most severe performance degradation ($\Delta$mIoU/Acc: -14.40 / -17.97), identifying it as the dominant component within our \emph{confidence-aware alignment mechanism}. Disabling the high-confidence gate $\tau_{\mathrm{high}}$ also results in a significant drop ($\Delta$mIoU/Acc: -5.49 / -7.98), which validates the necessity of adaptively selecting hard versus soft supervision at the region level. Finally, removing the CRF post-processing incurs a smaller but consistent decline ($\Delta$mIoU/Acc: -2.21 / -1.48), indicating its role as a complementary refinement rather than the primary driver of the gains. These trends hold consistently across other data splits. The full model produces visibly crisper boundaries, superior continuity for elongated structures, and fewer over-/under-segmentation artifacts, corroborating the quantitative improvements. \textit{Refer to supplementary material for more results.}

\vspace{1em}

\section{Conclusion and Future Work}
We present WakeupUrbanBench, the first benchmark dataset for semantic segmentation of historical Keyhole satellite imagery, and propose WakeupUSM, a self-supervised framework that combines frozen feature extractors with SAM-generated pre-pseudo-labels. Experiments show that WakeupUSM outperforms existing unsupervised methods on both binary and multi-class tasks.
Looking ahead, we identify several future directions to advance historical urban analytics: (1)  \textbf{Temporal expansion across decades.} Extending WakeupUrbanBench to the 1980s enables long-term urban morphology analysis. (2) \textbf{Cross-region generalization.} Adapting the framework to global cities supports cross-cultural comparisons in urban studies.

\newpage
{
    \small
    \bibliographystyle{ieeenat_fullname}
    \bibliography{main}
}


\end{document}